\documentclass[letterpaper, 10 pt, conference]{ieeeconf}

\IEEEoverridecommandlockouts                              

\usepackage{lipsum}
\usepackage{graphicx}                       
\usepackage{stfloats}
\usepackage[tight,footnotesize]{subfigure}  

\usepackage{amssymb,amsmath}
\usepackage{textcomp}
\usepackage{mdwmath}
\usepackage{mdwtab}
\usepackage{eqparbox}
\usepackage{rotating}                       
\usepackage{array}                          
\usepackage{listings}                       
\usepackage[ruled,vlined,noresetcount]{algorithm2e}
\usepackage{listings}
\usepackage[noend]{algpseudocode}

\usepackage[colorlinks,
            linkcolor=blue,
            anchorcolor=blue,
            citecolor=blue]{hyperref}

\usepackage{graphicx}
\usepackage{float}
\usepackage{subfigure}
\usepackage{amsmath}
\usepackage{algpseudocode}
\usepackage{stfloats}
\usepackage{multirow} 
\usepackage[ruled,vlined]{algorithm2e}
\usepackage{amssymb}
\usepackage{booktabs}

\title{\LARGE \bf An Integrated LiDAR-SLAM System for Complex Environment with Noisy Point Clouds}
\author{Kangcheng Liu$^{*}$, Aoran Xiao, Jiaxing Huang, Kaiwen Cui, Yun Xing, and Shijian Lu$^{*}$
}

\begin{document}

\maketitle
\thispagestyle{empty}
\pagestyle{empty}

\begin{abstract}
The current LiDAR SLAM (Simultaneous Localization and Mapping) system suffers greatly from low accuracy and limited robustness when faced with complicated circumstances. From our experiments, we find that current LiDAR SLAM systems have limited performance when the noise level in the obtained point clouds is large. Therefore, in this work, we propose a general framework to tackle the problem of denoising and loop closure for LiDAR SLAM in complex environments with many noises and outliers caused by reflective materials. Current approaches for point clouds denoising are mainly designed for small-scale point clouds and can not be extended to large-scale point clouds scenes. In this work, we firstly proposed a lightweight network for large-scale point clouds denoising. Subsequently, we have also designed an efficient loop closure network for place recognition in global optimization to improve the localization accuracy of the whole system. Finally, we have demonstrated by extensive experiments and benchmark studies that our method can have a significant boost on the localization accuracy of the LiDAR SLAM system when faced with noisy point clouds, with a marginal increase in computational cost. 
\end{abstract}
 \section{Introduction}
The 2D/3D simultaneous localization and mapping (SLAM) is of great significance to various applications, such as autonomous driving, 3D robotics grasping, search and rescue robots, and robotics inspection. The Light Detection And Ranging (LiDAR) sensor is the most popular and widely adopted 3D sensor for obtaining accurate depth information of all surrounding objects within a local range. Recently, various LiDAR SLAM systems have been proposed thanks to the development of large-scale LiDAR-based localization benchmarks such as KITTI \cite{behley2019semantickitti, geiger2012we} for autonomous driving.  Learning-based 3D vision has a great potential to leverage limited data to aid localization and perception \cite{liu2022enhance, liu2022Integrated, liu2017avoiding, liu2020fg, liu2019deep, zhao2021legacy, liu2022weakly, liu2022light, liu2022semi, liu2022robust}. 
 
 Currently, many low-cost LiDAR sensors have been developed such as the Livox-AVIA \cite{lin2020loam} for outdoor circumstances and the RealSense L515 for indoor scenarios. However, they suffer greatly from the noise when deployed into real circumstances. According to our experiments, the state-of-the-art LiDAR SLAM system suffers greatly from low accuracy when faced with real complicated circumstances. For example, according to our experiments detailed later in the leftmost of Fig. \ref{fig_outdoor}, we have done SLAM tests for the range of 500$m$ $\times$ 500$m$ outdoor garden scenario based on LEGO-LOAM \cite{shan2018lego, liu2022industrial, liu2022TIE, liu2022ws3d}. This scenario is with various building glass materials that reflect light, and the localization suffers greatly from the noise and outliers. It can be seen that the mapping performance is inferior because of the strong noise in obtained point clouds from LiDAR. 
\begin{figure}[t]
\setlength{\belowcaptionskip}{-0.68cm}
\centering
\includegraphics[scale=0.251006]{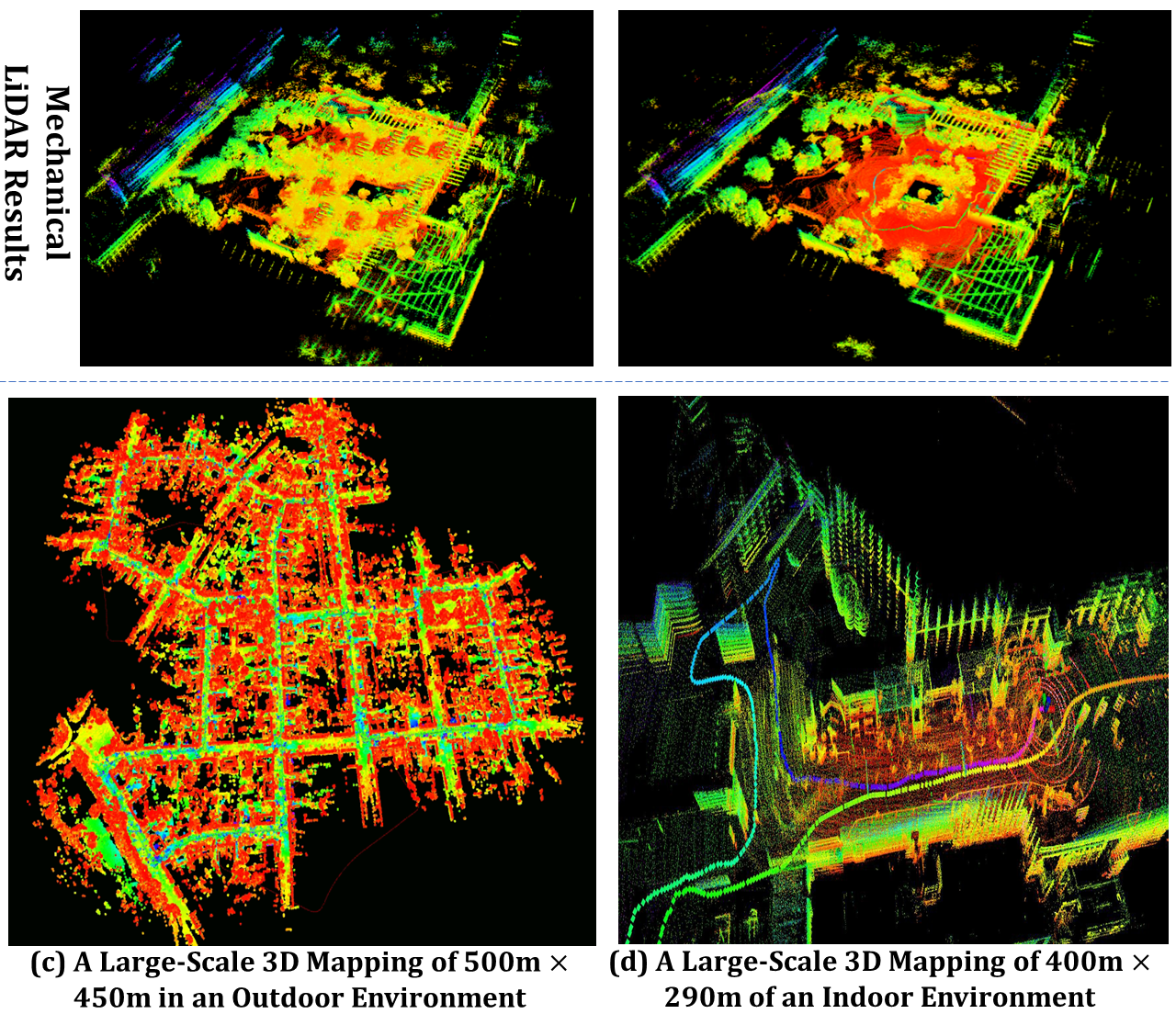}
\caption{Subfig. (a) shows the reconstructed map with noisy LiDAR points caused by reflective materials in complex environments. Subfig. (b) shows the mapping result by our proposed system. Our proposed LiDAR-inertial localization and mapping system is appropriate for both outdoor (c) and indoor (d) mapping with mechanical and solid-state LiDARs.}
\label{fig_overall}
\vspace{-4.99mm}
\end{figure}
 Also, the drift is very large when the motion of the ground vehicle system is very fast when doing the mapping. According to our experiments shown later in Fig. \ref{fig_outdoor}, in the circumstances with noisy point clouds, simply integrating State-of-the-art (SOTA) loop closure detection methods such as scan context \cite{kim2021scan} can not help the vehicle find loop closures in a noisy global map. Therefore, two major challenges remain for the LiDAR SLAM system to perform localization in complicated circumstances. The first is how to deal with the noises and outliers in the local map, especially for the SLAM of low-cost LiDAR sensors. The second is how to deal with the drifting problem, and to design an efficient place recognition approach for loop closure in large-scale complicated circumstances.

To tackle the challenges mentioned above, in this work, we have proposed \textit{D-LC-Nets}: robust denoising and loop closing networks for LiDAR SLAM in complicated circumstances with noisy point clouds. As shown in Fig \ref{fig_overall}, the robust, accurate, and consistent localization mapping can be realized by our system for both outdoor (Subfigure (b))/indoor (Subfigure (a)) circumstances. In summary, we have the following main contributions:

\begin{enumerate}
\item We have proposed a lightweight network for local map denoising in SLAM. We have proposed specific designs, including the two-scale CNN based feature extraction and RNN based feature fusion, which achieves accurate denoising. 


\item We have proposed a effective loop closure network to improve the localization accuracy and robustness. The classification and regression loss are both designed to extract the place-specific features and to find direction deviations of two LiDAR scans. The experiments demonstrate its effectiveness.

\item  We have integrated the proposed methods into our system both with solid-state LiDARs (Livox-AVIA) and mechanical LiDARs (VLP-16). Satisfactory performance is achieved for challenging benchmarks and real circumstances including gardens and corridors with strong light reflections from glasses. It demonstrates the clear merits of our methods in effectiveness and efficiency.

\end{enumerate}
This work is given as follows: Section II introduces related work and our overall system framework.  Section III introduces our proposed approach, including the denoising network for the local map (Denoise-Net), and the loop closure detection network (LC-Net). Finally, we have shown extensive experiments and briefly introduced our integrated system for on-site inspections.

\section{Related Work}
\subsection{LiDAR Localization and Mapping}
  LiDAR SLAM has been explored extensively in the past years. Cartographer \cite{hess2016real} proposed by Google utilizes the Ceres solver to solve the nonlinear least-square to achieve matching in scanning. The high-frequency visual information and low-frequency LiDAR odometry information are both utilized to achieve the robot motion estimation \cite{zhang2015visual}. The LEGO-LOAM was proposed in \cite{shan2018lego} for fine-grained feature extraction. And the real-time localization and mapping performance can be achieved by this system with acceptable accuracy in a simple environment. Recently, some methods have also been proposed to tackle the problem of limited field of views for solid-state LiDAR, such as Loam-Livox\cite{lin2020loam}. And LiDAR-Inertial SLAM systems have been proposed including the Fast-LIO\cite{xu2021fast}, which uses the incremental KD-Tree in the front-end of odometry. The iterative Extended Kalman Filter is utilized for the backend optimization. The deep learning based method has been proposed for frontend feature extraction recently \cite{sarlin2020superglue}. However, the major problem is that previous methods have not included or have a poor performance in loop closure and denoising in their system. Therefore, the LiDAR SLAM systems suffer from drifting and limited accuracy.

\subsection{Loop Closure and Denoising} 
The loop closure detection methods can be divided into vision-based and LiDAR-based approaches. A typical vision-based method is the bag of words model \cite{galvez2012bags}, which measures the smallest distance of visual words to the trained vision vocabulary. Superglue \cite{sarlin2020superglue} based learned feature has also been introduced, but it suffers from limited domain adaptation capacity, such as the transition between indoor and outdoor scenarios. The LPD-Net \cite{liu2019lpd} and OverlapNet \cite{chen2021overlapnet} are point-based networks for the place recognition, and the scan-context-based approach has also been proposed to perform loop closure.  However, their performance in real-world complicated circumstances is limited according to our experiments. \\

\section{Our System Architecture and Proposed Methodology}
\subsection{Our Overall System Framework}
\begin{figure}[htbp!]
\centering
\includegraphics[scale=0.227]{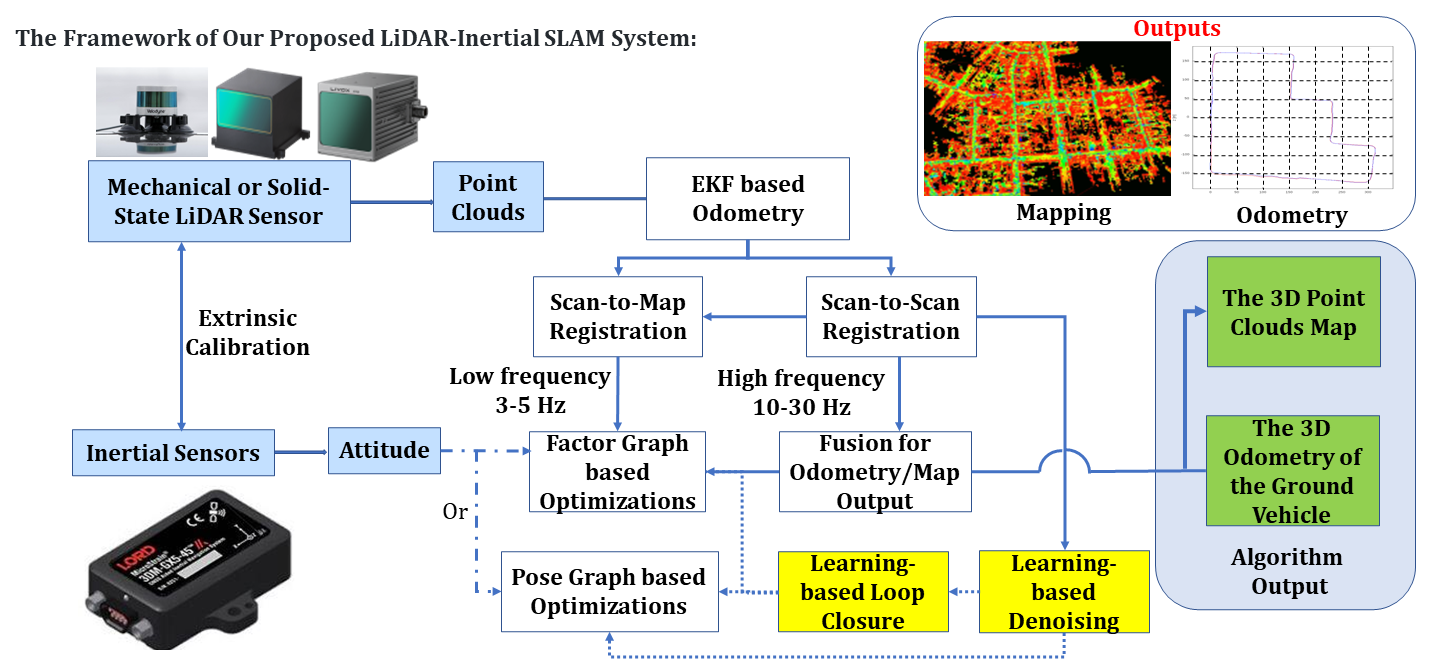}
\caption{The framework of our proposed LiDAR-Inertial localization and mapping system. Our proposed approach is appropriate for both the mechanical LiDARs and the solid-state LiDARs.}
\label{fig_frame}
\vspace{-2.9mm}
\end{figure}

Our proposed framework is mainly based on the typical LiDAR SLAM system for ground vehicles: LEGO-LOAM \cite{shan2018lego} and LIO-SAM \cite{shan2020lio}. Based on them, we have proposed two new modules to increase the overall localization accuracy. The first is that we propose an efficient network architecture for point clouds noise removal on the local map. The second is that we propose an effective loop closure detection network termed LC-Net. The proposed two networks can realize real-time inference based on the local map obtained from scan-to-scan registration in LiDAR SLAM. The overall accuracy of the whole LiDAR SLAM system is greatly improved by the proposed approaches. We follow scan-context \cite{kim2021scan} for the scan-to-scan matching based loop closure in place recognition. The directional deviation of two same LiDAR scans for loop closure is also used for the directional compensations.
\subsubsection{The Framework of Our Proposed LiDAR-Inertial Localization and Mapping System}
 The framework of our overall LiDAR SLAM system is summarized in Fig. \ref{fig_frame}. Our proposed approach takes the raw measurement from the LiDAR to obtain the point clouds and the raw measurement from the IMU sensors to obtain attitude. Then, we utilized the obtained point clouds to conduct the extended Kalman Filter optimization in the front end to construct the local map. Then, the scan-to-scan registration can be performed at a high frequency to obtain the connected local map, and the scan-to-map registration can be conducted at a lower frequency to construct the global map based on the connected local map. The fused output gives the dense 3D point clouds map and the 3D odometry of the ground vehicle simultaneously. Finally, we have supported both the factor graph based and the pose graph based backend optimization to improve the accuracy of the global map. Also, our proposed learning based loop closure detection strategy can correct the large drifts in the global map, and more consistent mapping of the environment can be achieved. Our proposed approach is both appropriate for the mechanical LiDAR such as the Velodyne VLP-16 and the solid-state LiDAR such as Livox-AVIA.
\subsubsection{The Final Software Framework for the Proposed LiDAR-Inertial SLAM System}

\begin{figure*}[ht]
\centering
\includegraphics[scale=0.11166]{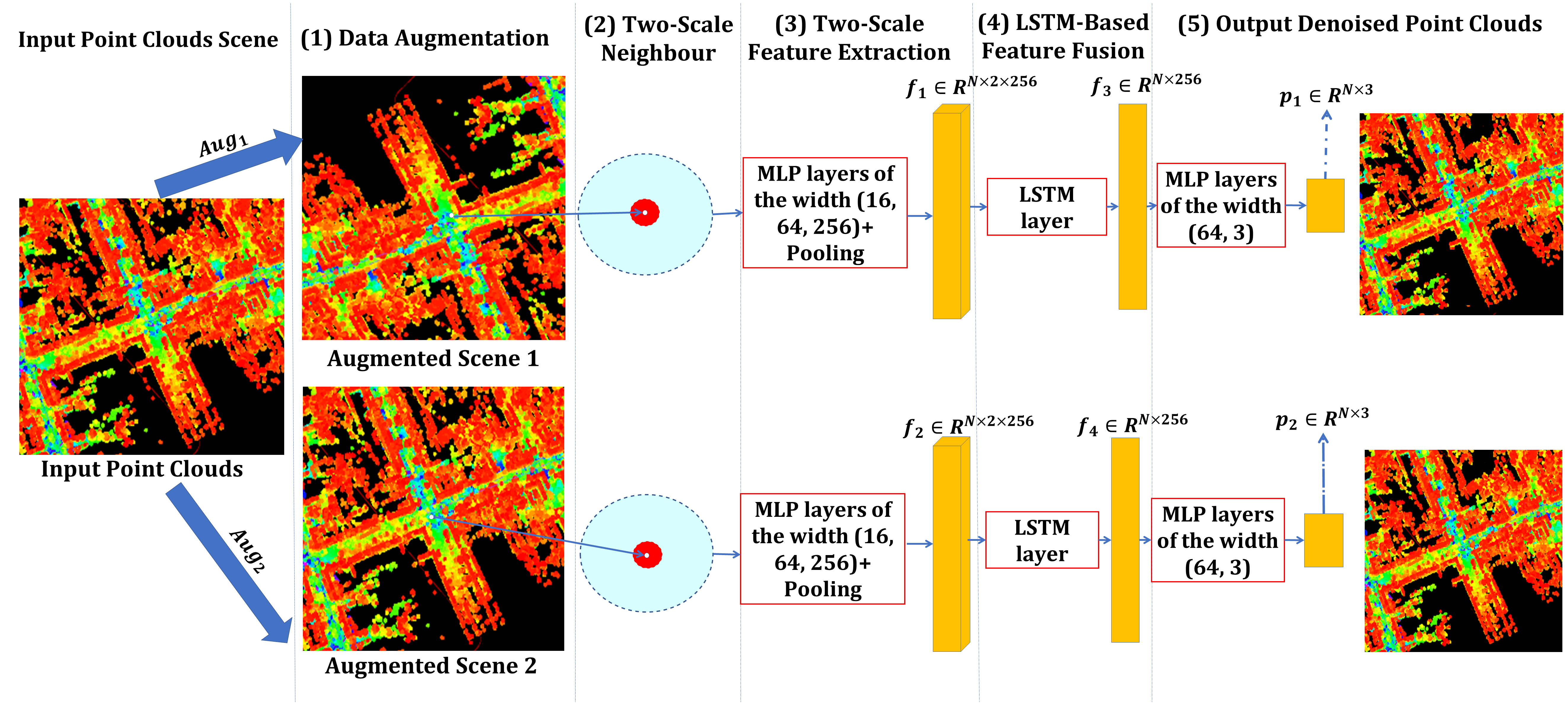}
\caption{The overall framework of our proposed local map denoising network (Denoise-Net). The network consists of five parts sequentially to do denoising.}
\label{fig_dnet}
\vspace{-5mm}
\end{figure*}

  Our software system is summarized as follows: 1. Laser motion distortion calibration. The pre-integration is calculated by using the IMU data between the start and end of the current frame. The pose of the laser point at each moment is obtained, to transform it into the coordinate system of the laser at the initial moment to realize the calibration. 2. Extract features. For the current frame laser point cloud after motion distortion correction, the curvature of each point is calculated, and then the corner and planar point features are extracted based on LIO-SAM \cite{shan2020lio}. 3. Scan-to-map matching. Extract the feature points of the local keyframe map, perform scan-to-map matching with the feature points of the current frame, and update the pose of the current frame.
4. Our proposed framework supports both pose graph and factor graph based optimization for the backend. We add laser odometry factor, GPS factor, and closed-loop factor to the graph to perform factor graph optimization, and update all keyframe poses.
5. Loop closure detection. We follow Scan-context \cite{kim2021scan} to find candidate loop closure matching frames in historical keyframes and perform scan-to-map matching to obtain the pose transformation. Then we can build loop closure factors, and add them to the factor graph for optimization.
Note that the only sensors we used are LiDAR and IMU. We include the GPS fusion module in our SLAM software and sensor suites but we have not utilized the GPS information in all our experiments.
\subsection{Efficient Network Architecture for Point clouds Noise Removal}
\subsubsection{Motivations}
We observe in real applications, the performance of the loop closure detection depends highly on the quality of obtained local point clouds map. And the accuracies of the loop closure and the overall localization performance are strongly dependent on the accurate and free-from-noises local point map. In this work, we have proposed a point cloud denoising approach for the obtained local point map. The noise removal network can be integrated into the SLAM system to obtain a relatively clean local map for the loop closure detection, which improves the overall success rate of loop closure detections and finally improves the localization accuracy.
\subsubsection{Network Framework}
The overall framework of our proposed local point clouds denoising network (D-Net) is shown in Fig. \ref{fig_dnet}. Improved on \cite{behley2015efficient}, we proposed three conditions for a faster KNN query. \textbf{1}. If the octant is not intersected with the query ball, we should skip it. \textbf{2}. If the query ball is within the octant, we stop searching. \textbf{3}. If the query ball includes the octant, searching the children of that octant is not required. We greatly improve the KNN query speed by 18.2 times. The network architecture is proposed as follows in five steps: (1). Firstly, we do data augmentations for the input point clouds with rotation, scaling, and perturbation, to obtain the augmented point clouds in local scene 1 and local scene 2. (2). Secondly, the points are sampled in the neighborhood of two different scales of the local map. To be more specific, we randomly sample 16 points in the red region, which is the small scale; and we also randomly sample 16 neighbor points in the blue region, which is the large scale. (3). Then we adopt MLP based feature extraction for the two scales respectively. We have utilized three layers of MLP, with a width of 16 layers, 64 layers, and 256 layers, respectively. And subsequently, we apply max-pooling to the feature obtained from the 16 neighbour points to obtain the features $f_1, f_2 \in \mathbb{R}^{N\times 2 \times 256}$  after MLP layers for the two augmented scenes, respectively. The two denotes the features are of two diverse scales and 256 denotes the dimension. (4) Then, we propose to utilize the LSTM \cite{HochSchm97} based network layers to adaptively fuse features from the large and small scales, and reduce the scale-level dimension to one. Because from our observations, besides spatial correlations in the obtained point clouds from two different scales by the LiDAR scanners, temporal correlations also exist. We utilize the LSTM layer to model the temporal correlations between the two different scales. (5) We use MLP layers to obtain final denoised point clouds and use proposed losses to guide the optimization process.

\subsubsection{Optimization Loss Functions}
Loss functions are proposed to acquire high-quality denoised point clouds with clean and distinctive features for loop closure detection. It is demonstrated by our experiments that by our proposed optimization function considering both denoising and normal estimation, we can obtain clean denoised results with a uniform and reasonable distribution of point clouds. Note that all our proposed losses are merely required in training. \\
\textbf{Chamfer Distance based Loss}
Chamfer Distance based Loss has recently been demonstrated to be effective in point clouds related low-level tasks. To force denoised point clouds $\mathop{\textbf{P}}\limits^{*}$ to be consistent with ground truth point clouds $\textbf{P}^{\mbox{\textit{\tiny{GT}}}}$, we have designed an adapted Chamfer Distance based loss for the point clouds denoising task. However, in constructing the local maps, we observe that for the edges and corners of surrounding objects that are very important in feature extractions, point clouds obtained by the LiDAR sensor are often sparse. While for the planar regions that are very smooth, the obtained point clouds by the LiDAR sensor are usually very dense. The typical Chamfer Distance \cite{wu2021density} based loss for computer vision can not deal well with large-scale local maps. The fine-grained geometric details will be overlooked by the network because of the too small amount of corner or edge points are considered in training. We have proposed a geometric-aware Chamfer Distance based loss based on the local curvature as follows:

We have utilized the efficient off-the-shelf method \cite{zhang2008curvature} to calculate the average curvature of local point clouds. Because the transformation between the point clouds' surface curvature and surface normal is very easy, we can also obtain the surface normal of the local point clouds. Denote the normalized curvature value of the ground truth point clouds as $\sigma$. Then we can obtain the geometric-aware Chamfer Distance based loss $L_{Geo}$ denoted as follow:
\begin{equation}
\begin{aligned}
        L_{Geo} =&\frac{1}{|\mathop{\textbf{P}}\limits^{*}|}\sum_{p_i \in \textbf{P}^{\mbox{\textit{\tiny{GT}}}}}\sigma\mathop{min}\limits_{{\mathop{p}\limits^{*}}_i \in \mathop{\textbf{P}}\limits^{*} }(1-\exp^{-\|p_i-{\mathop{p}\limits^{*}}_i\|}) + \\
        &\frac{1}{|\textbf{P}^{\mbox{\textit{\tiny{GT}}}}|}\sum_{p_i \in \mathop{\textbf{P}}\limits^{*}}\sigma\mathop{min}\limits_{p_i \in \textbf{P}^{GT}}(1-\exp^{-\|{\mathop{p}\limits^{*}}_i-p_i\|}).
\end{aligned}
\end{equation}
Where $p_i$ is the \textit{i}-th point of the ground truth point clouds $\mathop{\textbf{P}}^{\mbox{\textit{\tiny{GT}}}}$, ${\mathop{p}\limits^{*}}_i$ is the \textit{i}-th point of the predicted denoised point clouds $\mathop{\textbf{P}}\limits^{*}$. The local curvature $\sigma$ gives larger weights to the edge and the corner of the local point clouds and gives smaller weights to smooth surfaces. \\
\textbf{Loss for the Local Distribution} To encourage the point clouds after denoising to have a uniform and consistent distribution, we have also proposed the following repulsive loss function:
\begin{equation}
\begin{aligned}
        L_{R} =\sum_{i}\sum_{j \in K(i,j)}exp\{ -\lambda \Vert{\mathop{p}\limits^{*}}_{i}- {{{\mathop{p}\limits^{*}}}_{i,j}}\Vert^2\}
\end{aligned}
\end{equation}

Where $K(i, j)$ is the \textit{j}-th index set of the $K$-nearest neighbour of the center query point. This loss will encourage the point pairs not to be too adjacent to each other. Thus, the denoised point clouds can maintain a uniform and reasonable distribution. \\
\textbf{Loss of the Normal Regulation}
The normal is a very significant feature for the 3D point clouds. In our proposed network, we have utilized the normal as an evaluation of the denoising quality. To be more specific, we have simply designed a normal regulation loss to make the normal of the predicted denoised point clouds consistent with the ground truth. To be more specific, denote the network predicted normal vector and the ground truth normal vector as $\textbf{n}_i$ and $\textbf{n}^{gt}_i$ respectively. we simply encourage the normal of predicted denoised point clouds $\textbf{n}_i$ consistent with the ground truth normal vector $\textbf{n}^{gt}_i$. The loss function design is formulated as follows:
\begin{equation}
\begin{aligned}
        L_{N} =\sum_{i}\Vert{\textbf{n}_i-\textbf{n}^{gt}_i}\Vert^2
\end{aligned}
\end{equation}
Finally, the final loss function $L_{den}$ for local map denoising is the sum of the three losses proposed above, which can be given as:
\begin{equation}
\begin{aligned}
        L_{den} =L_{Geo}+L_{R}+L_{N}
\end{aligned}
\end{equation}
\textbf{Training Setting of Proposed Network} 
Our proposed network is trained in an end-to-end manner with a deep learning framework \textit{Pytorch}. We train our network with a single NVIDIA RTX 2070 GPU for 180 epochs with the Adam optimizer, utilizing a learning rate of $10^{-4}$ with a decay of 0.1 per 60 epochs. Because our proposed approach is aimed at large-scale LiDAR SLAM, we train our denoise-Net at the training set of large-scale real-world denoising benchmark \textit{Paris-rue-Madame} \cite{serna2014paris}. And we directly do inference in the real-world experiments and the benchmarks. The training takes approximately 16 hours. Note that all our proposed loss function designs are merely required in the training stage. 

\begin{figure}[tbp!]
\centering
\includegraphics[scale=0.063]{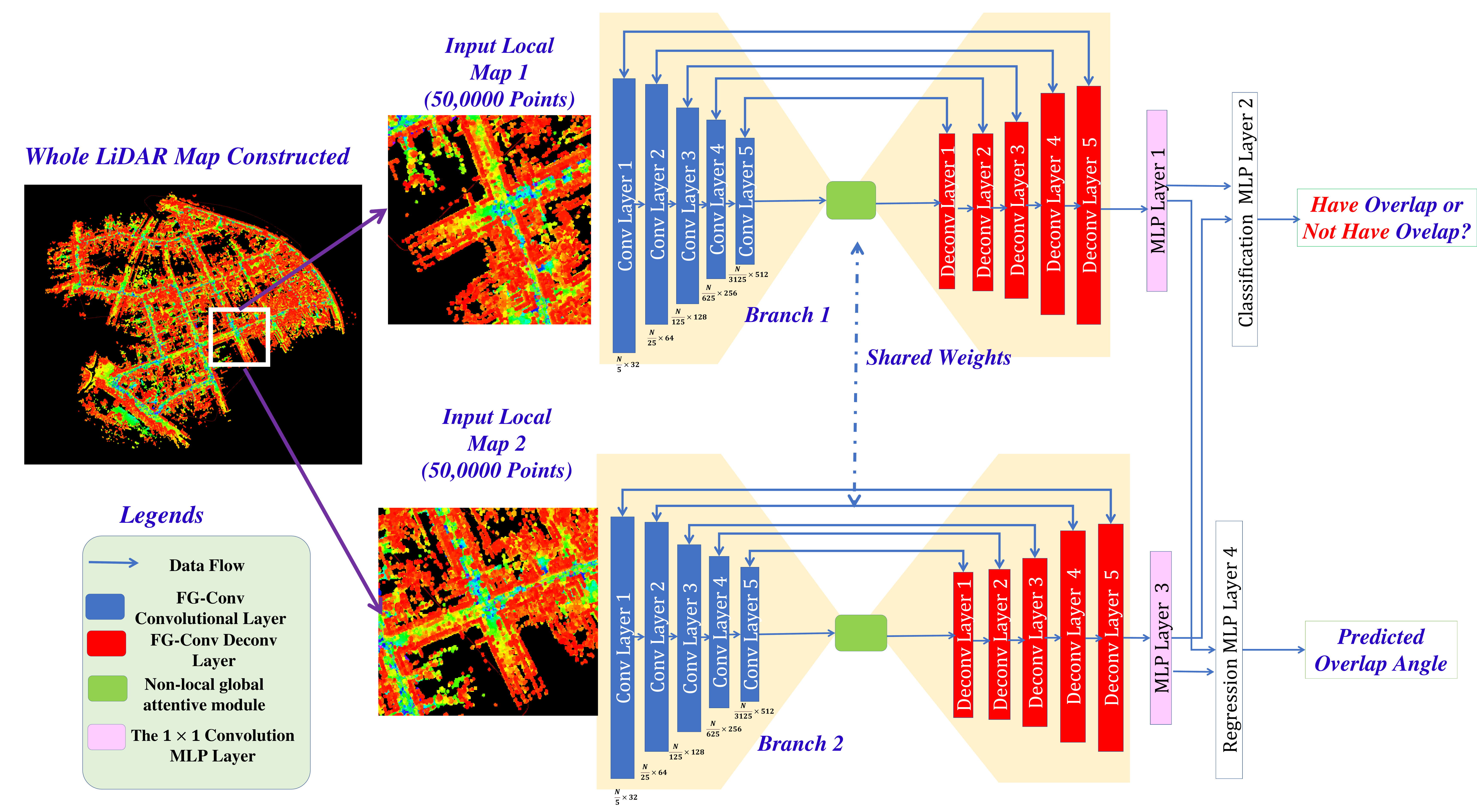}
\caption{The overall framework of our proposed loop closure network (LC-Net). We propose to input the LiDAR point clouds to the encoder-decoder based network to give the prediction that if the two LiDAR scans have overlap between each other. Also, we will regress the overlap angle between those two scans to facilitate the subsequent global map correction and the back-end optimization.}
\label{fig_loop_clo}
\vspace{-4.97mm}
\end{figure}





\subsection{The Learning based Loop Closure}

The loop closure network is based on our previous work FG-Conv \cite{liu2021fg, liu2022weaklabel3d, liu2022fg}, as shown in Fig. \ref{fig_loop_clo}. In our setting of the training of the network, we have adopted a learning rate of $5 \times 10^{-4}$ with a decay of 0.98 in each training epoch. And the training of our network lasts for 280 epochs with the Adam optimizer. We implement the network framework using \textit{Pytorch}.
 \section{Experimental Results}
\begin{figure*}[tbp!]
\centering
\includegraphics[scale=0.418]{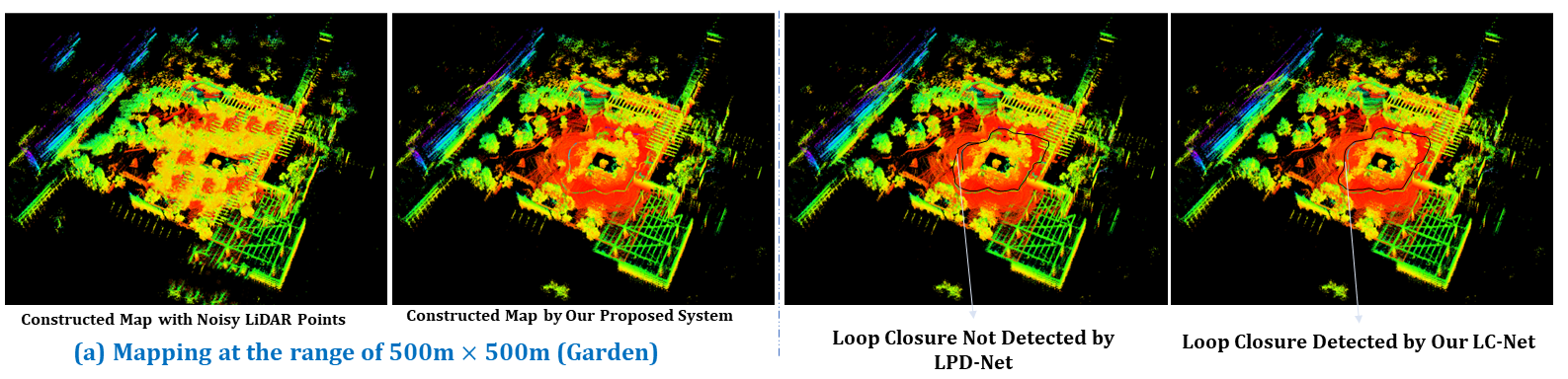}
\caption{Real experiment results of the proposed approach. We have done UGV SLAM with mechanical LiDAR (Subfig. (a) Above) in the garden and Solid-State LiDAR (Subfig. (b) below) within the corridors of buildings. The left shows the comparisons of noise removal results, and the right shows the comparisons before, and after our loop closure detection approach. It can be demonstrated that with our proposed noise removal approach, a more consistent global map can be achieved, and the loop closures in the ground vehicle can be effectively detected. It demonstrates that our proposed approaches have improved the overall accuracy and robustness of the LiDAR SLAM system in complex environments equipped with various LiDAR sensors.}
\label{fig_outdoor}
\vspace{-5.8mm}
\end{figure*}
 \subsection{Experimental Setup}
 In this section, we present comprehensive experimental results about the performance of SLAM with the proposed approach under various circumstances. We have deployed our proposed denoising and loop closing network onto NVIDIA TX2 onboard computer. For test case 1, we have utilized the UGV mounted with mechanical LiDAR Velodyne VLP-16 and Inter Realsense D435 for our experiments. For test case 2, we have utilized the Unmanned Ground Vehicle (UGV) mounted with solid-state LiDAR Livox-AVIA and Inter Realsense D435 for our experiments. The first reason we tested with solid-state LiDAR is that we want to show that our proposed approaches are also appropriate for solid-state LiDAR. The second reason is that we want to reduce the influences from the reflection of the building glass by the properties of the solid-state LiDAR. Also, we have tested under various complicated circumstances as shown in Fig. \ref{fig_outdoor}. 
 For the benchmark testing, we have also done extensive experiments on the KITTI benchmark for the performance evaluation of noise removal and loop closure. We have compared the performance of noise removal and loop closure for the quality of the global map in detail.
\subsection{Experimental on the Unmanned Ground Vehicle}
\subsubsection{Experimental Setup}
 We have done experiments with mechanical LiDAR and with solid-state LiDAR. As shown in Fig. \ref{fig_outdoor}, for the first test case, we utilized the UGV to simulate a commodity transportation task in the garden environment with mechanical Velodyne VLP-16 LiDAR. For the second test case, we utilized the UGV to perform a surveillance task with solid-state LiDAR Livox AVIA within the corridor. The left shows the comparisons of noise removal results, and the right shows the comparisons before, and after our loop closure detection approach. It can be demonstrated that with our proposed noise removal approach, a more consistent global map can be achieved. For the loop closure, we have also integrated the current State-of-the-art loop closure detection method LPD-Net \cite{liu2019lpd} to the UGV SLAM system for comparisons. It can be seen that the LPD-Net fails to detect the loop closure in complicated garden circumstances with trees, plants, buildings, etc. And in the narrow corridor case below with various strong reflections in the building glass, the loop closure can not be detected. While our proposed denoise-net can handle various complicated backgrounds, the loop closures in the ground vehicle can be effectively detected.  It demonstrates that our proposed approaches have improved the overall accuracy and robustness of the LiDAR SLAM system in complex environments equipped with various LiDAR sensors.

 \subsubsection{Performance Comparisons} We have also recorded the localization accuracy compared with the ground truth and the success rate in loop closure detection of the tested case 1 of the garden environment in Fig. \ref{fig_outdoor} Subfig. (a), and the cooridor environment among buildings shown in Fig. \ref{fig_outdoor} Subfig. (b). As shown in Table \ref{table_table}, the global localization accuracy can be significantly improved by our proposed learning based loop closure detection method. When tested without our proposed noise removal network, the localization error is doubled. And without our proposed loop closure approaches, the localization error is even larger. If the computational time is less than 100 ms per scan (more than 10 Hz), the real-time localization and mapping can be achieved. Note that the inference time of proposed approach only has marginal increase on SOTAs and we can realize the SLAM task in real-time with more than 10 Hz, which demonstrated the efficiency of our proposed approaches. As shown in Table  \ref{table_loop_close}, with our proposed learning based loop closure detection approach, the success rate of the loop closure is significantly improved compared with State-of-the-art approaches. Thus, the robustness of the whole SLAM system is guaranteed.

 \begin{table}[t]
\caption{The comparisons of localization accuracy and computational time for every capture LiDAR scan tested at case one garden environment shown in Fig. \ref{fig_outdoor} in the range of around 500m $\times$ 500m (Left Value), and the case two of corridor environment of around 350m $\times$ 400m (Right value). In all the experiments, ours (Full) denotes our full SLAM framework with both \textit{D-Net} and \textit{LC-Net}. Ours (w/o Denoising) denotes our framework without \textit{D-Net}. Ours (w/o Loop) denotes our framework without \textit{LC-Net}. }
\label{table_table}
\begin{center}
\resizebox{\linewidth}{!}{\begin{tabular}{lll}
\toprule
Methods & Computation Time (ms / frame) & Error (cm) / 200 m\\
\hline
Ours (w/o Loop)&74.58 / 76.37&12.35 / 12.29\\
Ours (w/o Denoising)&71.62 / 75.29&12.17 / 12.35\\
Ours (Full)&82.36 / 85.25 &11.98 / 11.82\\
LEGO-LOAM \cite{shan2018lego}&48.9 / 46.63&13.65 / 13.68\\ 
LIO-SAM \cite{shan2020lio}&47.5 / 49.58 &12.56 / 12.63\\
\bottomrule
\end{tabular}}
\end{center}
\vspace{-4.8mm}
\end{table}

\begin{table}[tbp!]
\caption{The Comparisons Results of the Loop Closure Test}
\label{table_loop_close}
\begin{center}
\resizebox{\linewidth}{!}{\begin{tabular}{lcc}
\toprule
Methods & Success & Failure in Loop Closure\\
\hline
LPD-Net (w/ \textit{D-Net} Denoising) \cite{liu2019lpd}&5&2\\
LPD-Net (w/o \textit{D-Net} Denoising) \cite{liu2019lpd}&4&3\\
Ours (w/o Denoising)&5&2\\
Ours (Full)&7&0\\
LEGO-LOAM \cite{shan2018lego}&1&6\\
LIO-SAM \cite{shan2020lio}&3&4\\
\bottomrule
\end{tabular}}
\end{center}
\vspace{-4.5mm}
\end{table}
\begin{table}[tbp!]
\caption{The Comparison Results of the Denoising Approaches on the Localization Performance with the LiDAR SLAM experiments in the Garden Environment (Left Value) and the Corridor Environment (Right Value) shown in Fig. \ref{fig_outdoor}.}
\label{table_denoise}
\begin{center}
\resizebox{\linewidth}{!}{\begin{tabular}{lc}
\toprule
Methods in Comparisons & Average Localization Error (cm)\\
\hline
Ours (Full)& 11.98 / 11.82\\
Ours (w/o $L_{Geo}$)& 12.02 / 11.91\\
Ours (w/o $L_{R}$)& 12.05 / 12.03\\
Ours (w/o $L_{N}$)& 12.13 / 12.21\\
Ours (w/o $L_{ce}$)& 12.19 / 12.25\\
Ours (w/o $L_{reg}$)& 12.18 / 12.16\\
GDCNN \cite{wei2021geodualcnn} & 12.92 / 12.89\\
SPCD \cite{luo2021score} &12.25 / 12.29\\
Point-Clean \cite{rakotosaona2020pointcleannet}&12.51 / 12.58\\
PCGLR \cite{zeng20193d}&12.67 / 12.73\\ 
\bottomrule
\end{tabular}}
\end{center}
\vspace{-6mm}
\end{table}

\begin{figure}[htbp!]
\centering
\includegraphics[scale=0.236]{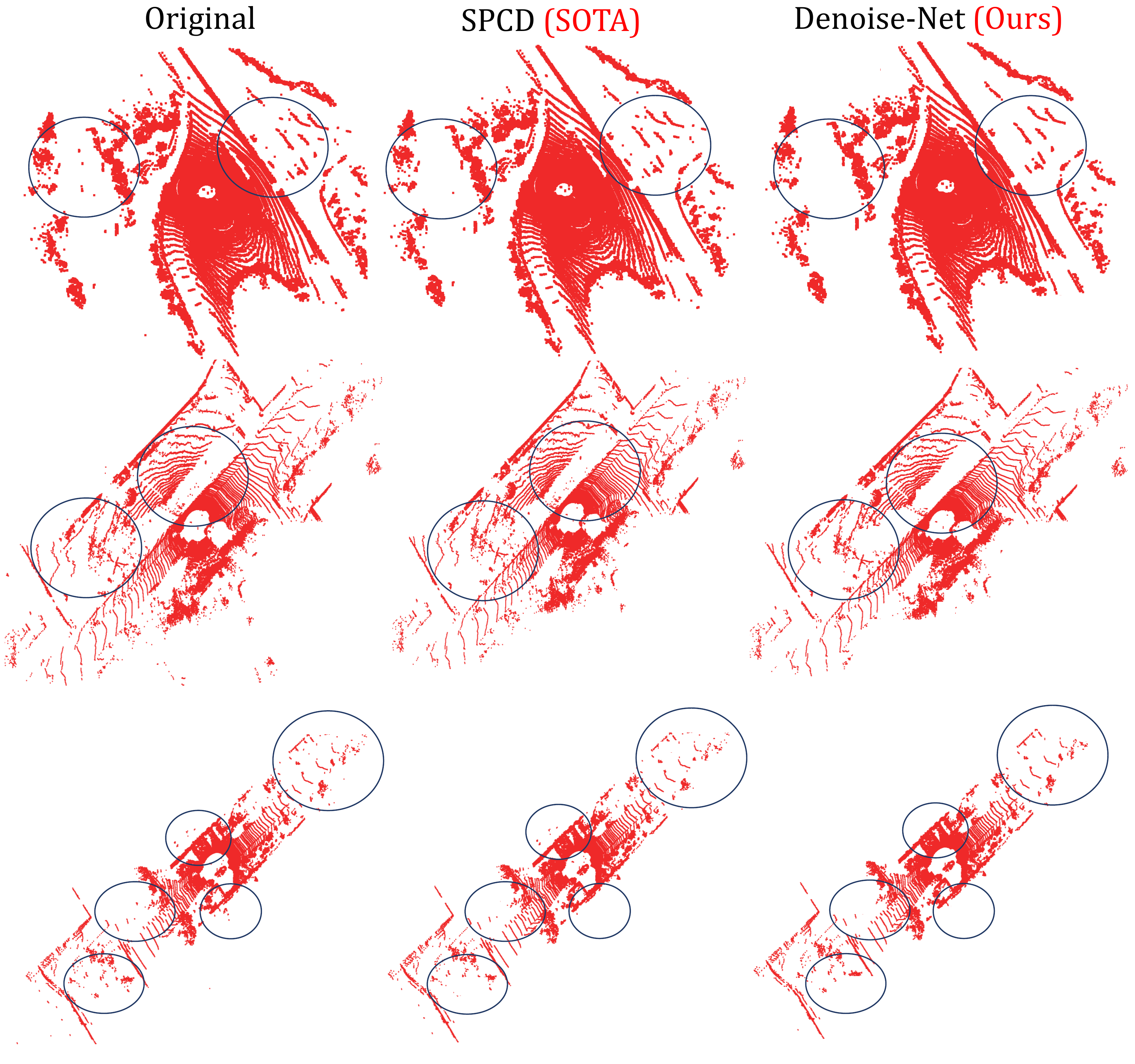}
\caption{The final experiments of the noise removal on the KITTI public benchmark. The detailed denoising comparisons are highlighted in black circles. Compared with existing SOTA methods SPCD \cite{luo2021score}, we can provide qualitatively and quantitatively more consistent and clean denoising results, which is of great significance to the subsequent loop closure and mapping. And real experiments have demonstrated the effectiveness of our approach in real-world circumstances.}
\label{fig_noise}
\vspace{-1.1mm}
\end{figure}

\begin{figure}[htbp!]
\centering
\includegraphics[scale=0.26]{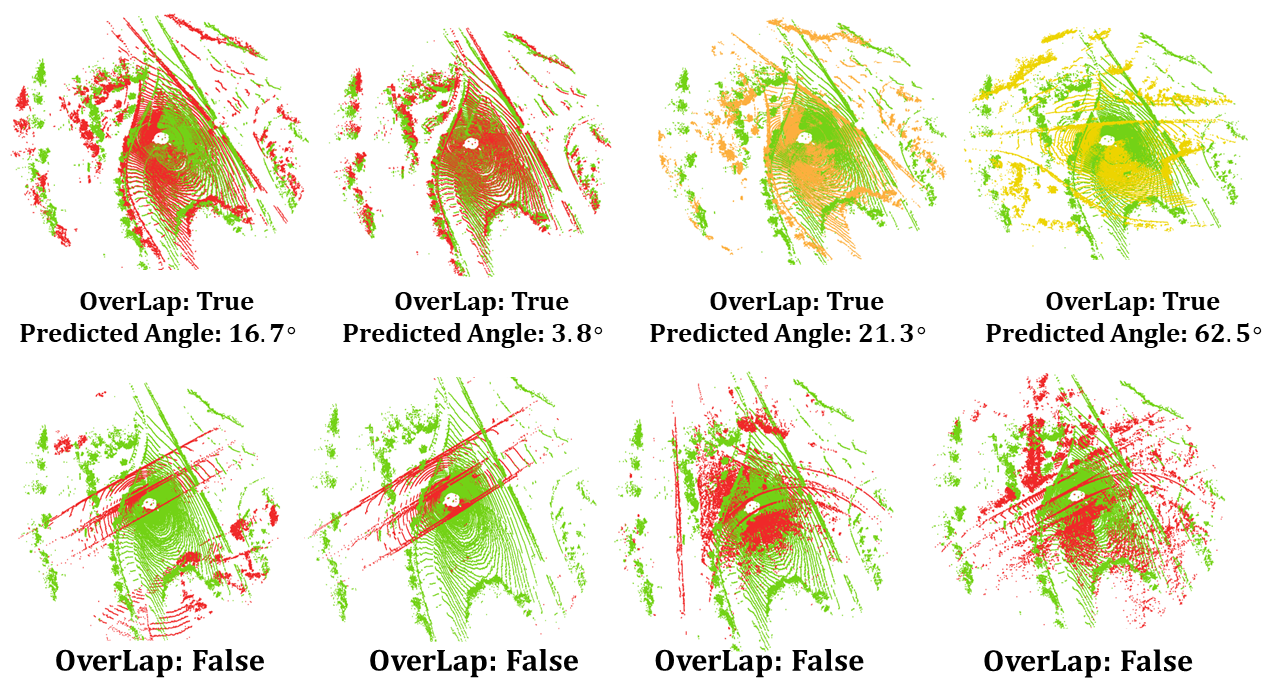}
\caption{The demos of the prediction of our proposed network for the loop closure of the crossroad. If two different LiDAR scans have large overlaps with each other which means the overlap is true, it indicates the two point clouds scans belong to the same scan and align well with each other. If the overlap is small, it means the two point cloud scans belong to two different scans. It can be seen from the first row that our proposed framework can accurately detect the same scan, which means the two scans have overlap with each other. Also, the relative angle predicted is very accurate. In addition, from the second row, it can be seen that our network can accurately differentiate between the two different LiDAR scans. }
\label{fig_close}
\vspace{-2mm}
\end{figure}



\begin{figure}[ht]
\centering
\includegraphics[scale=0.268]{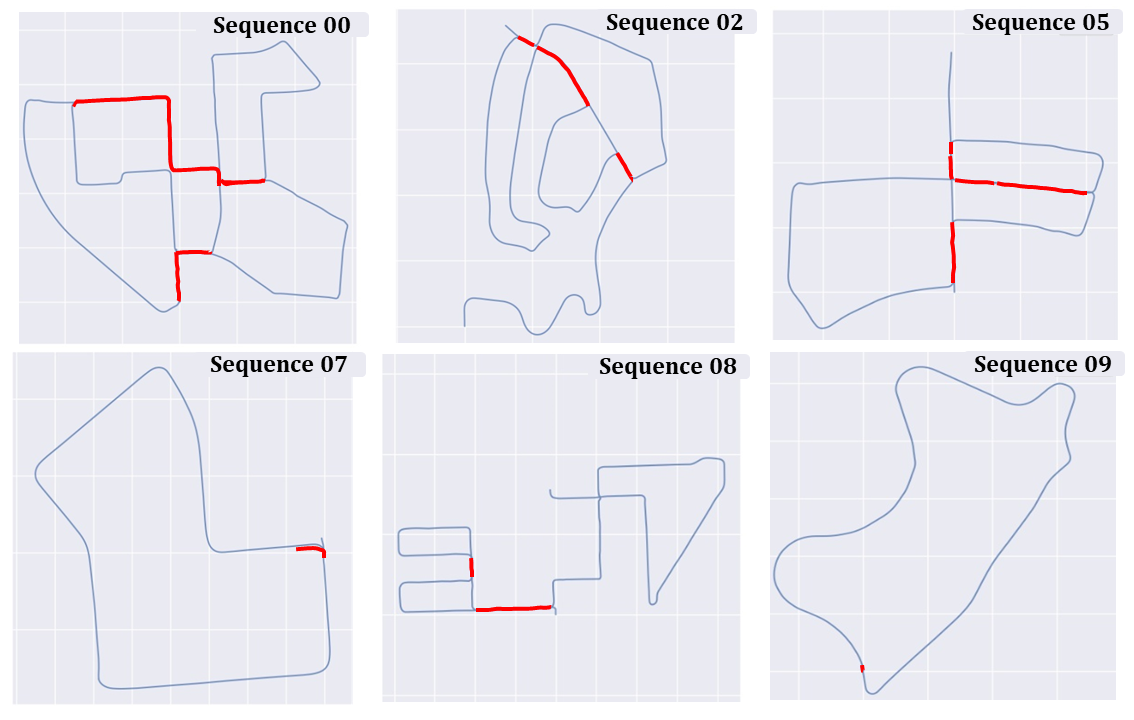}
\caption{Experiments of the place recognition on the KITTI public benchmark. The loop closures detected by our loop closing network is marked with the red color. We have tested all the KITTI sequences with apparent loops. It is demonstrated that our proposed approach which is termed as LC-Net, has high precision in the place recognition for loop closure. And further experiments has demonstrated the effectiveness of our approach in real-world circumstances.}
\label{fig_seq}
\vspace{-1.3mm}
\end{figure}
 \subsection{Experiments on Public Benchmarks}
To demonstrate the robustness and effectiveness of our proposed approach in the outdoor large-scale autonomous driving circumstances. The sequences of the KITTI dataset is also collected with the Velodyne HDL-64 LiDAR, and is challenging for finding the loop closure due to the dynamic monotonous environments. 
 \subsubsection{Testing of the performance of the Denoise-Net}
Firstly, we have tested the performance of our proposed Denoise-Net, the results are shown in Fig. \ref{fig_noise}. Compared with existing SOTA methods SPCD \cite{luo2021score}, we can provide qualitatively and quantitatively more consistent and clean denoising results, which is of great significance to the subsequent loop closure and mapping. And experiments in the garden environment have demonstrated the effectiveness of our approach in real-world circumstances, as shown in Table. \ref{table_denoise}. We have done detailed ablation studies of proposed network losses and results are shown from rows 1-6 in Table. \ref{table_denoise}. Various denoising approaches vary significantly in inference time and many can not realize real-time denoising. It can be demonstrated all proposed losses improve the overall localization accuracy. To directly test the performance of various approaches, we have also tested offline for the performance of the localization accuracy in the garden environment. Also, it can be seen that our proposed approach, termed Denoise-net, has performed minimum global localization error. It can be interpreted by the fact that the design networks for point clouds or mesh denoising are merely appropriate for the analysis of small-scale objects with fine-grained details, but not appropriate for the large-scale localization problem which requires explicit edges and corner features for feature extraction and feature matching. 
  
  \subsubsection{Testing of the performance of the LC-Net}
We have also tested the performance of our proposed LC-Net, the results are shown in Fig. \ref{fig_close}, and Fig. \ref{fig_seq}. We have separated the KITTI benchmark into single scans of different directions in the same sequence. From the first row of Fig. \ref{fig_close}, it can be demonstrated that our proposed network can precisely detect the same scan that revisits the same place in a different direction. From the second row of Fig. \ref{fig_close}, it can be seen that our proposed framework can accurately distinguish between two different scans. It demonstrates that our proposed LC-Net is very effective in loop closure detection. We have also tested the proposed LC-Net for the place recognition task. The results for the sequences 00, 02, 05, 07, 08, and 09 are shown. Fig. \ref{fig_seq} shows the loop closure detection result. The trajectories obtained from GPS are shown in black, and the detected loop closing places are marked in red. The quantitative results of the prevision and recall rate on KITTI sequences 00, 02, 05, 08 are shown in Table \ref{table_SOTA_kITTI}. The precision rate indicates the ratio of correctly paired scans, and it can avoid mistakes in the scan-to-map matching. The recall rate represents the percentage of correctly recorded loop closures, which is important in correcting the drifts in the global map. We have also included the vision-based approach with the bag of words \cite{galvez2012bags} for comparisons, which is widely adopted in current vision-based SLAM systems. Although the vision-based method \cite{galvez2012bags} can achieve high precision (100\% for sequence 02 and sequence 05), the recall is relatively low compared with our proposed LC-Net. This is caused by the revisit the same place in the opposite direction. And the visual features can not be easily captured in this case. In addition, as demonstrated in Table \ref{table_SOTA_kITTI}, the effectiveness of our approach is merely comparable to or slightly better than other loop closure approaches for LiDAR SLAM when the denoise-net is not included. When the denoise-net is added, our proposed approach demonstrates more robust performance compared with SOTA methods, thanks to the complementary effects of our denoising and loop closure detection approach in improving the localization accuracy. 
\begin{table}[tbp!]
\caption{The Comparisons of Precision and Recall with SOTA Methods on KITTI Sequence 00/02/05/08, respectively.}
\label{table_SOTA_kITTI}
\begin{center}
\scalebox{0.8}{\begin{tabular}{lll}
\toprule
Methods & Precision (\%) & Recall (\%)\\
\hline
Ours (Full)&  \textbf{100.00} / 99.27 / 99.85 / \textbf{96.35}& \textbf{93.35} / \textbf{93.28} / \textbf{96.72} / \textbf{89.25}\\
Ours (w/o Denoise)&  93.25 / 95.63 / 97.68 / 93.80& 92.36 / 92.53 / 92.77 / 86.12\\
LPDNet \cite{liu2019lpd}& 91.65 / 93.47 / 93.62 / 91.57&82.10 / 90.26 / 88.91 / 86.25\\
BOW \cite{galvez2012bags}& 100.00 /  \textbf{100.00} /  \textbf{100.00} / 90.25 &92.00 / 80.60 / 87.60 / 79.50\\
S-Context \cite{kim2021scan}& 100 / 90.00 / 100.00 / 92.50&87.00 / 73.00 / 90.00 / 76.60\\ 
O-Net \cite{chen2021overlapnet}&98.00 / 98.50 / 96.90 / 90.65&89.05 / 88.70 / 89.30 / 86.50\\
\bottomrule
\end{tabular}}
\end{center}
\vspace{-2.16mm}
\end{table}
\section{Conclusions}
In this work, we have proposed an integrated improved LiDAR-Inertial simultaneous localization and mapping system for unmanned aerial/ground vehicles. To improve the localization accuracy, we have proposed a noise removal method for the constructed local map. Subsequently, we have proposed systematical designs of the learning-based loop closure detection method to improve the accuracy of the final global map. It is demonstrated by extensive experiments that our proposed LiDAR-Inertial SLAM system shows great accuracy and robustness in various circumstances with various testing devices, including both indoor and outdoor circumstances with mechanical and solid-state LiDAR. Also, real-time performance for mapping at more than 10Hz in large complicated environments can be achieved by our proposed system.

\addtolength{\textheight}{0cm}   





\bibliographystyle{IEEEtran}
\bibliography{references}

\end{document}